\documentclass[conference]{IEEEtran}
\IEEEoverridecommandlockouts
\usepackage{cite}
\usepackage{amsmath,amssymb,amsfonts}
\usepackage{multirow}
\usepackage{algorithmic}
\usepackage{enumitem}
\usepackage{hyperref}
\usepackage{tablefootnote}
\usepackage{float}

\usepackage{graphicx}
\usepackage{svg}
\svgsetup{inkscapepath=.}
\usepackage{textcomp}
\usepackage{tikz}
\usetikzlibrary{arrows.meta}
\usetikzlibrary{positioning, shapes.geometric, calc, fit, backgrounds}
\usepackage{xcolor}
\def\BibTeX{{\rm B\kern-.05em{\sc i\kern-.025em b}\kern-.08em
    T\kern-.1667em\lower.7ex\hbox{E}\kern-.125emX}}
\begin{document}

\title{Image Quality Assessment of Identity Cards Using Measures from Open Face Image Quality
}

\author{\IEEEauthorblockN{1\textsuperscript{st} Gregor Grote, 2\textsuperscript{nd} Juan E. Tapia, 3\textsuperscript{rd} Christian Rathgeb}
\IEEEauthorblockA{\textit{da/sec - Biometrics and cybersecurity research group} \\
\textit{Darmstadt University of applied science, Darmstadt, Germany }\\
gregor.grote, juan.tapia-farias, christian.rathgeb\{@h-da.de\}}
}
\maketitle

\begin{abstract}

This paper addresses the challenge of assessing image quality in ID cards in remote verification systems by applying capture-related quality measures from the Open Face Image Quality (OFIQ) standard to ID card images.
Our preprocessing pipeline includes corner detection, perspective normalization, and comprehensive foreground masking to ensure accurate and unbiased quality measure computation.
We evaluate the effectiveness of these measures by analyzing their correlation with the performance of three presentation attack detection (PAD) algorithms across four diverse ID card datasets, where two datasets contain bona fide, i.e. pristine, images and two contain printed mock ID cards.
Our results suggest that quality assessment based on some OFIQ measures can significantly improve PAD performance.

\end{abstract}


\section{Introduction}

Remote verification of identity documents (ID) has become an increasingly important topic in recent years, especially during and after the COVID-19 pandemic \cite{GomezBarrero-BiometricsCOVID-TTS-2022}.
ID cards are widely used for identity verification in remote contexts, such as access to online banking and government services.
To ensure reliable identity verification and prevent fraud, PAD systems rely on the quality of ID card images.
Low-quality images can lead to rejection of legitimate documents or even misclassifications of presentation attacks.
Therefore, developing effective methods for assessing the quality of ID card images is essential for enhancing security and reliability of remote identity verification systems.

Image quality assessment (IQA) methods should be explainable to the user and the system operator to understand the reasons behind the quality score and take appropriate actions to improve image quality if necessary.
The current standard for assessing the quality of biometric face images \cite{ISOIEC29794-5-2025} is implemented in the Open Face Image Quality (OFIQ) \cite{BSI_OFIQ1_0_2024} framework.
However, there is no standardized approach for assessing the quality of ID card images.
Quality assessment of ID card images is challenging due to the wide variety of ID card designs, materials, and security features.
For example, OFIQ determines whether the illumination of the face image is uniform by comparing the illumination of patches located at the subject's cheeks, as they usually have a similar skin tone and are located symmetrically on the face. Such an approach is not applicable for ID cards as there are generally no such reference points on ID cards.
Furthermore, developing PAD systems for ID cards is challenging because of privacy concerns, as there are no public datasets available that contain images of real ID cards \cite{pad_overview}.

\begin{figure}[t]
    \centering
    \begin{tikzpicture}[
        scale=0.85, transform shape,
        node distance=1.0cm and 0.8cm, 
        every node/.style={font=\sffamily\footnotesize, align=center},
        process/.style={rectangle, draw, fill=blue!5, inner sep=5pt, minimum width=3cm, minimum height=0.8cm},
        decision/.style={diamond, draw, fill=orange!5, aspect=1.8, inner sep=1pt, minimum width=2.5cm},
        terminal/.style={rectangle, draw, fill=gray!10, minimum width=2.5cm, minimum height=0.7cm},
        arrow/.style={->, >=stealth, thick},
        labelnode/.style={font=\sffamily\scriptsize\itshape},
        imgnode/.style={inner sep=0pt}
    ]

    \node[process] (acq) {Data Acquisition};

    \node[below left=0.5cm and -0.8cm of acq, imgnode] (badimg)
        {\includegraphics[width=2.2cm]{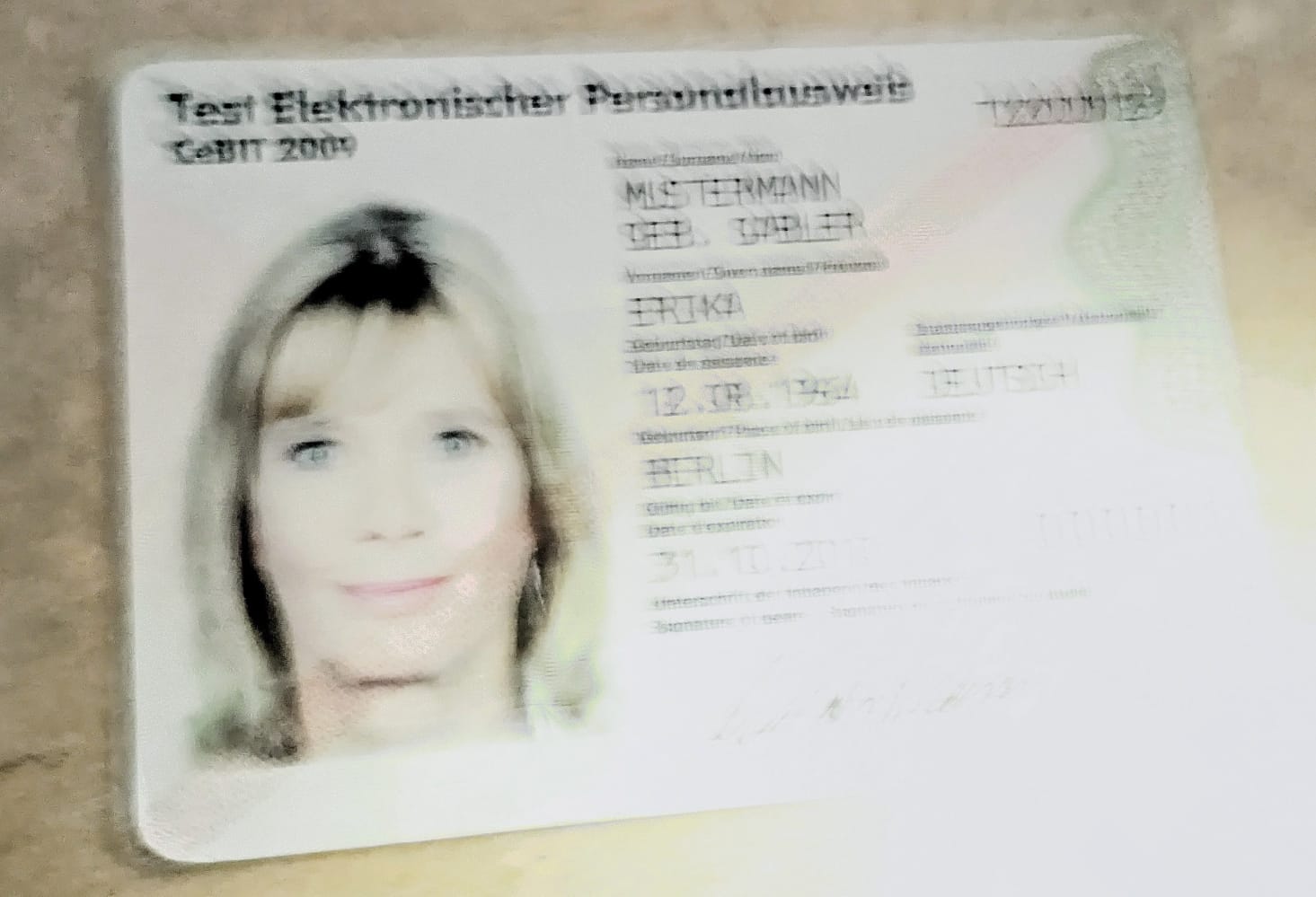}};
    \node[below right=0.5cm and -0.8cm of acq, imgnode] (goodimg)
        {\includegraphics[width=2.2cm]{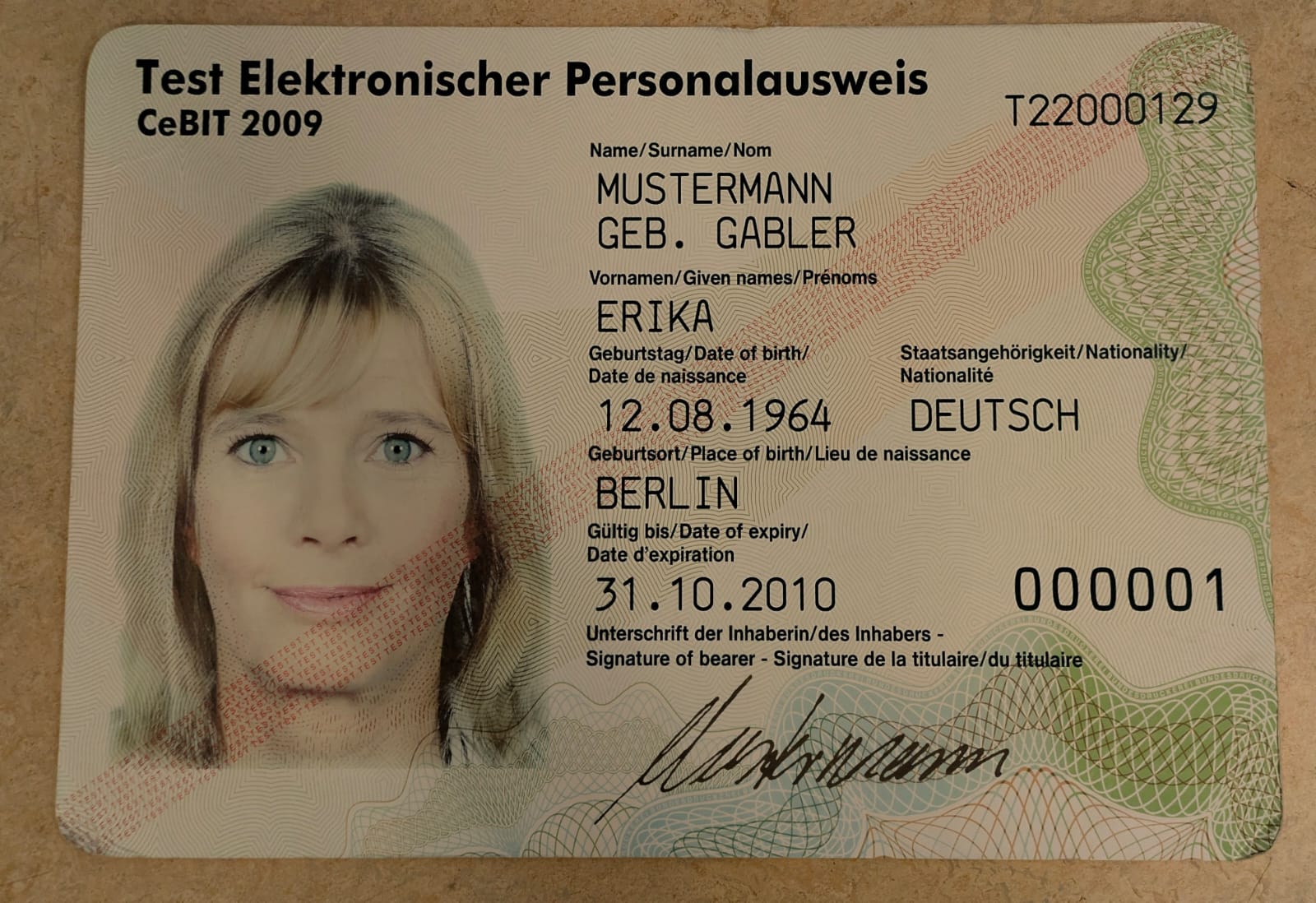}};
    
    \node[below=0.1cm of badimg, labelnode] (lab1) {(a) Low Quality Sample};
    \node[below=0.1cm of goodimg, labelnode] (lab2) {(b) High Quality Sample};

    \node[process, below=3.2cm of acq, text width=3.2cm] (pre) {
        \textbf{Preprocessing Pipeline}\\
        \tiny (Corner Detection, Normalization, Masking)
    };

    \node[process, below=0.7cm of pre, text width=3.2cm] (quala) {
        \textbf{Quality assessment}\\
        \tiny (Based on OFIQ)
    };

    \node[decision, below=0.7cm of quala] (iqa) {Quality sufficient?};

    \node[process, below left=0.7cm and 0.1cm of iqa, draw=red!70, fill=red!5, text width=3cm] (discard) {
        \textbf{Discard \& Feedback}\\
        \scriptsize Reject: Insufficient Quality
    };

    \node[process, below right=0.7cm and 0.1cm of iqa] (pad) {PAD System};


    
    \draw[arrow] (acq.south) -- +(0,-0.25) -| (badimg.north);
    \draw[arrow] (acq.south) -- +(0,-0.25) -| (goodimg.north);

    \draw[arrow] (lab1.south) -- +(0,-0.15) -| ([xshift=-0.8cm]pre.north);
    \draw[arrow] (lab2.south) -- +(0,-0.15) -| ([xshift=0.8cm]pre.north);

    \draw[arrow] (pre.south) -- (quala.north);
    \draw[arrow] (quala.south) -- (iqa.north);

    \draw[arrow] (iqa.west) -| (discard.north) 
        node[midway, above] {No};
    
    \draw[arrow] (iqa.east) -| (pad.north) 
        node[midway, above] {Yes};



    \end{tikzpicture}
    \caption{Workflow of quality assessment in remote identification system. Low-quality captures are discarded, while acceptable images pass through to the PAD system.}
    \label{fig:qa_workflow}
\end{figure}
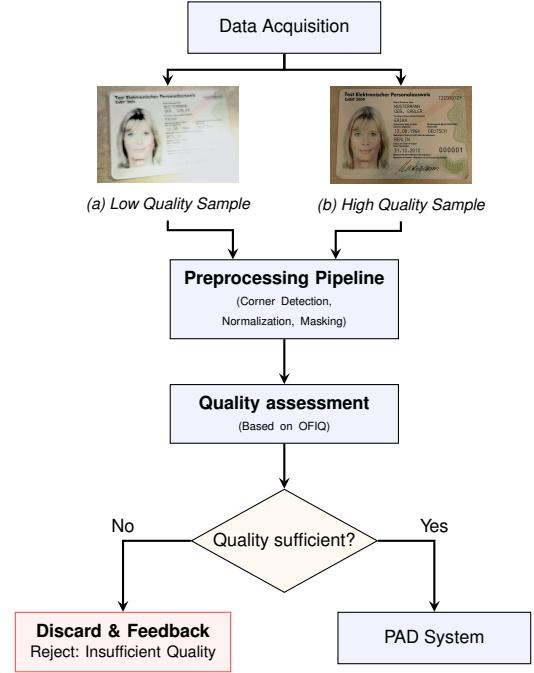

This work aims to develop and evaluate image quality assessment methods specifically for ID card images by adapting quality measures from OFIQ to ID card images.
This could help to improve the performance of PAD systems on ID card images, if low-quality images that may lead to misclassifications with PAD systems are discarded, see Figure~\ref{fig:qa_workflow}.

In recent years, many research efforts have been devoted to face image quality assessment (FIQA) \cite{Schlett-survey}, while only some recent works have focused on IQA for ID cards \cite{pad_overview}. Yañez et al. \cite{quality_brisque} proposed an IQA system for Chilean ID cards. The authors analyze multiple image features to predict subjective scores. A dataset of 204 ID card images was evaluated by 15 subjects, enabling the assessment of model performance. Schulz et al. \cite{quality_ocr} presented another approach for IQA for ID cards. A key contribution of their work is a hybrid framework leveraging both FIQA on the face image and text-based quality measures to improve PAD prediction accuracy in a private dataset. Gonzalez et al. \cite{Gonzalez} investigated the influence of FIQA measures on the detection of presentation attacks, including printed, screen-displayed, and composite ID card images from a private dataset. 
Al-Ghadi et al. \cite{IdTrust} analyzed a latent representation of the ID card background to distinguish between bona fide images and pictures of scans, as scans usually do not capture fine details in the background. However, this latent representation is not interpretable for an operator.

Motivated by these prior challenges, we propose and adapt methods to evaluate the impact of OFIQ measures on ID cards rather than only on faces. 

The main contributions of this work are:
\begin{itemize}
    \item Adaptation of OFIQ capture-related measures for ID card images.
    \item Development of a new algorithm for the illumination uniformity measure tailored for ID card images.  
    \item Evaluation of the effect of IQA for ID cards on PAD performance.
\end{itemize}

This paper is organized as follows: IQA methods used for ID cards are described in detail in Section~\ref{sec:methods}. Used datasets and experimental results are summarized in Section~\ref{sec:dbs} and ~\ref{sec:results}, respectively. Finally, Section~\ref{sec:conclusion} concludes and provides an outlook to future research.

\section{Methods}\label{sec:methods}

We used quality measures from OFIQ \cite{BSI_OFIQ1_0_2024} and reimplemented them for ID card images.
OFIQ uses a set of quality measures that are divided into two categories: capture- and subject-related quality measures.
Capture-related quality measures are related to the image acquisition process and include measures such as illumination and sharpness.
In contrast, subject-related quality measures are related to the facial characteristics of the subject in the image and include measures such as pose, expression, and facial occlusion.
To adapt OFIQ for ID card images, only capture-related quality measures were used. 
In total, OFIQ uses 10 capture-related quality measures.
OFIQ also proposed two more measures (luminance skewness and luminance kurtosis) that were discarded in the final version because they did not improve face recognition performance. However, we include them in our analysis. A radial distortion measure was also proposed but was discarded because there is no algorithm to compute it yet.
We analyzed each of these measures, including the ones discarded by OFIQ (luminance skewness, luminance kurtosis, and radial distortion), and determined whether they can be directly applied to ID card images, need to be adapted, or are not applicable at all.
An overview of the measures and their applicability to ID card images is shown in Table \ref{tab:ofiq_metrics_application}.

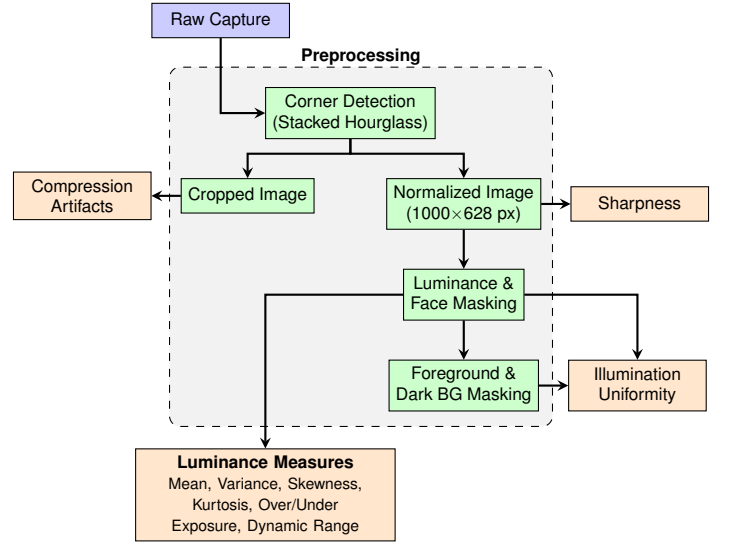
\begin{figure}[ht]
    \centering
    \begin{tikzpicture}[
        scale=0.65, transform shape,
        node distance=0.5cm and 0.4cm,
        every node/.style={font=\sffamily\normalsize, align=center},
        raw/.style={rectangle, draw, fill=blue!20, inner sep=5pt, minimum width=2.8cm, minimum height=0.7cm},
        prep/.style={rectangle, draw, fill=green!20, inner sep=4pt, minimum width=2.4cm, minimum height=0.7cm},
        metric/.style={rectangle, draw, fill=orange!20, inner sep=4pt, minimum width=2.8cm, minimum height=0.7cm},
        container/.style={draw, dashed, inner xsep=0.15cm, inner ysep=0.2cm, fill=gray!10, rounded corners},
        arrow/.style={->, >=stealth, thick}
    ]

    \node[raw] (ri) {Raw Capture};
    
    \node[prep, below right=1cm and -0.5cm of ri] (idc) {Corner Detection \\ (Stacked Hourglass)};
    
    \coordinate[above=0.1cm of idc] (topspacer);
    
    \node[prep, below left=0.8cm and -1.0cm of idc] (ci) {Cropped Image};
    \node[prep, below right=0.8cm and -1.0cm of idc] (ni) {Normalized Image \\ (1000$\times$628 px)};

    \node[prep, below=0.8cm of ni] (lum) {Luminance \& \\ Face Masking};
    
    \node[prep, below=0.8cm of lum] (mask) {Foreground \& \\ Dark BG Masking};

    \node[metric, left=0.6cm of ci] (ca) {Compression\\Artifacts};
    
    \node[metric, right=0.6cm of ni] (sharp) {Sharpness};
    
    \node[metric, right=0.6cm of mask] (ill_unif) {Illumination\\Uniformity};

    \node[metric, below left=6.3cm and -2.65cm of idc, text width=5cm] (ilumets) {
        \textbf{Luminance Measures} \\
        \small Mean, Variance, Skewness, Kurtosis, Over/Under Exposure, Dynamic Range
    };

    \draw[arrow] (ri) |- (idc.west);
    
    \draw[arrow] (idc.south) -- +(0,-0.3) -| (ci.north);
    \draw[arrow] (idc.south) -- +(0,-0.3) -| (ni.north);
    
    \draw[arrow] (ci) -- (ca);
    \draw[arrow] (ni) -- (sharp);
    \draw[arrow] (ni) -- (lum);
    \draw[arrow] (lum) -- (mask);
    
    \draw[arrow] (mask) -- (ill_unif);
    \draw[arrow] (lum.east) -| (ill_unif);
    \draw[arrow] (lum.west) -| (ilumets.north);

    \begin{scope}[on background layer]
        \node[container, fit=(topspacer) (idc) (ci) (ni) (lum) (mask)] (prep_group) {};
        
        \node[anchor=north, font=\sffamily\bfseries\normalsize, yshift=0.5cm] at (prep_group.north) {Preprocessing};
    \end{scope}

    \end{tikzpicture}
    \caption{Workflow of the image quality assessment (IQA) system. Preprocessing steps (green) generate the specific representations required for diverse quality measures (orange).}
    \label{fig:workflow}
\end{figure}

We decided not to use the background uniformity measure, as the background of ID card images is cropped prior to quality assessment.
The natural color measure is also not applicable, as ID cards can have different colors depending on the design of the card.
The radial distortion measure is not applicable, as ID cards are flat and therefore do not exhibit strong radial distortion.
Estimation of illumination uniformity requires a new algorithm, as the current algorithm compares the illumination of the cheeks in face images, which is not applicable for ID cards (see Subsection~\ref{sec:qa} for details).
For all other measures, the same algorithms as in OFIQ were used. However, to prevent biases in luminance-related measures, pixels belonging to faces were masked out. Figure \ref{fig:workflow} depicts a detailed workflow graph of the IQA employed for ID cards.

OFIQ also provides a conversion from the native quality measures values to a final quality score in the range of $0$ to $100$.
We decided not to use this conversion, as it is specifically designed for face images and may not be applicable for ID card images. 

\begin{table}
    \caption{Overview of OFIQ capture-related quality measures and if they are used for ID card image quality assessment.}
    
    \centering
    \begin{tabular}{|l|l|}
        \hline
        \textbf{Measure} & \textbf{Used}\\
        \hline
        Background uniformity      & No \\
        Illumination uniformity    & New algorithm \\
        Luminance mean             & Yes \\
        Luminance variance         & Yes \\
        Luminance skewness\tablefootnote{Discarded by OFIQ. Note that both OFIQ and we use the absolute value of skewness.}         & Yes \\
        Luminance kurtosis\tablefootnote{Discarded by OFIQ.}         & Yes \\
        Over exposure   & Yes \\
        Under exposure  & Yes \\
        Dynamic range              & Yes \\
        Sharpness                  & Yes \\
        Compression artifacts   & Yes \\
        Natural color              & No \\
        Radial distortion\tablefootnote{Proposed but no algorithm available in OFIQ.}          & No \\
        \hline
    \end{tabular}
    \label{tab:ofiq_metrics_application}
\end{table}

\subsection{Preprocessing}

Before applying the IQA measures, the ID card images were preprocessed to ensure that they were in a suitable format for analysis.
The first step was to detect the corners of the ID card. 
This was done using a stacked hourglass model \cite{stacked_hourglass} (see Section \ref{sec:dbs} for training details).
For inference, the input image was padded to a square aspect ratio, then an additional 10\% padding was added. The model is able to predict corner positions even when they are outside the image or occluded.
After that, the image was resized to $256\times256$ pixels, transformed to grayscale, normalized with a mean of $0.5$ and a standard deviation of $0.5$, and fed into the model.
The model returns four heatmaps, one for each corner of the ID card.
Subsequently, the pixel with the highest value was identified in each heatmap as the predicted corner position if this value exceeds $0.5$. Otherwise, the position was estimated using all other known corner positions.
When a corner is detected outside the image, the position is projected back to the closest point within the image.
Note that our current approach assumes an ID card is present in the image because the used datasets contain only images with ID cards.
In a real-world scenario, one would discard images if not all four heatmaps have peaks of at least $0.5$ and all peaks are within the image.



Using the predicted corners, a perspective transformation was applied to obtain a top-down view of the ID card with a resolution of $1000\times628$ pixels.
This step is referred to as normalizing the ID card.
This normalization helps to reduce the variability in the images due to different angles and distances from the camera. 
It also removes most of the background of the image, which is not relevant for the quality assessment.

Compression artifact detection was not performed on the normalized image, as potential compression artifacts from the image capture process may be altered during the normalization step.
Instead, the minimum bounding box around the detected corners was determined and used to crop the raw image.
We call this the "cropped image".

To determine the measures related to illumination and luminance, we compute the luminance $Y$ of each pixel using the same algorithm as OFIQ \cite{BSI_OFIQ1_0_2024}.

\begin{figure}[H]
    \centering
    \includesvg[width=0.4\textwidth]{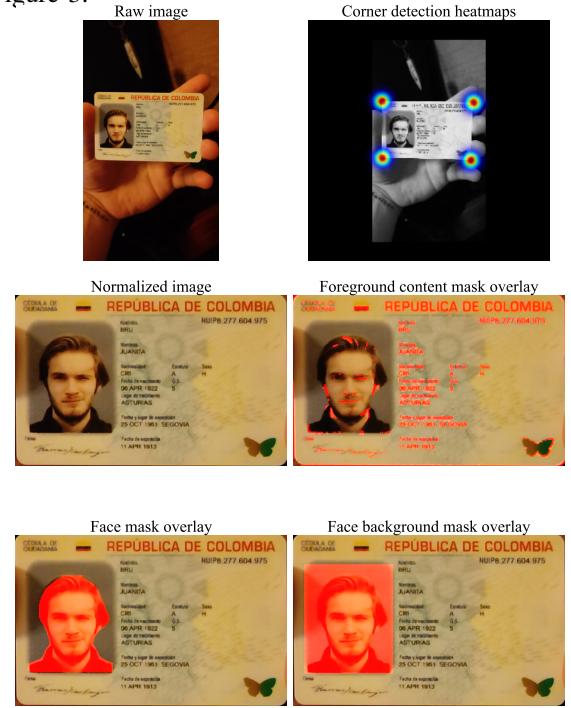}
    \caption{Example of preprocessing steps.}
    \label{fig:preprocessing}
\end{figure}

To prevent biases, all pixels that are part of face images on the ID card are masked out.
Face detection was done by the YOLOv11 medium segmentation model \cite{yolov11}.
Certain ID cards, for example some Colombian ID cards (see Figure \ref{fig:preprocessing}), were found to have a dark background behind the face image, while most other ID cards have a transparent background.
To compensate for this, a preprocessing step masks dark background pixels behind the face images.
To identify the dark background pixels, we compare the luminance of the bounding box of each face detected by YOLOv11 (excluding the face itself) with the rest of the image minus all faces and the foreground content from \cite{content_segmenter}.
If this area is significantly darker than the rest of the image, it is assumed that there is a dark background.
The bounding box is then expanded in small steps until the new pixels are no longer significantly darker than the rest of the image.

Furthermore, we avoid including text when determining the illumination uniformity measure, as a region with a lot of text could be wrongly interpreted as a dark shadow.
We tried different optical character recognition (OCR) methods to detect text and decided to use the EAST text detector \cite{east-ocr}, as it provided the best results on the employed datasets.
Using the text bounding boxes from the EAST text detector, the character pixels were masked out by combining a global threshold (within each bounding box) selected using Otsu's method \cite{otsu} and a local adaptive threshold.
However, the EAST text detector sometimes misses single characters and signatures.
Some ID cards, such as certain French ID cards, have barcodes, which cannot be detected by barcode detectors when the image is blurry.
Therefore, in a more general approach, we employed the algorithm from \cite{content_segmenter} to segment foreground content from the background.
This includes text, barcodes, and small coats of arms or logos on the ID card.
The false positives that the algorithm produces on face images, as in Figure \ref{fig:preprocessing}, can be ignored as the face will also be masked. 
We also tried bandpass filtering like in \cite{IdTrust}, but we found the algorithm from \cite{content_segmenter} to be more consistent.
Afterwards, we combine these three masks (face, dark face background, foreground content) to obtain a final mask of everything that is not considered background for the illumination uniformity measure.
An example of the preprocessing steps is depicted in Figure \ref{fig:preprocessing}.
\vspace{-0.5cm}

\begin{table*}
    \caption{Overview of used ID card datasets.}
    \centering
    \begin{tabular}{|l|l|l|l|l|l|}
        
        \hline
        \textbf{Dataset} & \textbf{Documents} & \textbf{Image count} & \textbf{Attacks} & \textbf{Bona fide-type} & \textbf{Availability} \\
        \hline
        CHI & Chilean ID cards & 40,963 & Composite, Print, Screen & Real & Private \\
        MEX & Mexican ID cards & 3,809 & Composite, Print, Screen & Real & Private \\
        IJCB PAD ID card 2025 track 1 (IJCB) \cite{IJCB2025_PAD} & Various ID cards & 12,000 & Composite, Print, Screen & Mock & Public \\
        ID card subset of DLC2021 \cite{dlc2021} & Various ID cards & 15,083 & Print, Screen & Mock & Public \\
        \hline
    \end{tabular}
    \label{tab:id_card_datasets}
\end{table*}

\subsection{Quality Assessment}\label{sec:qa}

After preprocessing, the quality measures from Table \ref{tab:ofiq_metrics_application} were computed on the preprocessed ID card images.
The implementation of the quality measures is identical to that used in OFIQ \cite{BSI_OFIQ1_0_2024}, except for the illumination uniformity measure, for which a new algorithm was developed.
First, the ID card image is divided into a grid of $3\times 4$ equally sized blocks, the foreground is masked out, and the median luminance of each block is computed.
Then the standard deviation of the median luminance values of all blocks is calculated as a measure of illumination uniformity.
A low standard deviation indicates that all $12$ regions have similar median luminance, while a high standard deviation could indicate a glare or dark shadow.

Figure \ref{fig:workflow} shows which quality measures require which preprocessing steps.
Open-source code for preprocessing and evaluation, including the pretrained models, can be found on \href{https://github.com/dasec/ID-Card-image-quality-on-OFIQ}{GitHub}.

\subsection{Analysis of Quality Measures}

To measure the effectiveness of the quality measures, their correlation with the performance of PAD systems on ID card images was analyzed.
The following pre-trained PAD systems were used:

\begin{itemize}
    \item The baseline model from track 2 (IJCB Baseline \cite{IJCB2025_PAD})
    \item Two private off-the-shelf PAD systems (PAD M1 and M2)
\end{itemize}

We hypothesize that higher quality images will lead to better PAD performance, while lower quality images may result in increased false positives or false negatives.
As the native scores from the individual quality measures were used, not all of them have a higher is better semantic.
For example, for the compression artifacts measure, a lower score indicates better quality.

\section{Datasets}\label{sec:dbs}

The employed datasets can be divided into two categories: private datasets with real bona fide images (CHI and MEX) and two public datasets containing printed mock ID card images (IJCB and DLC2021) and their respective attacks.
The bona fide images are images of real ID cards, while printed mock ID card images are images of custom printed plastic ID cards, which are then captured.
Printed mock images are easier to obtain because they do not require access to real ID cards that contain sensitive personal information.
While a PAD system trained on printed mock images may perform well on other datasets with printed mock images, it may not generalize well to real bona fide images due to a possible domain gap between printed mock and real bona fide images.

As PAD M1 and PAD M2 were trained on real bona fide images, the printed mock images from IJCB and DLC2021 were labeled as attacks for these systems.
The IJCB baseline model was trained on mock bona fide images and identified $2$\% of mock bona fide images from IJCB and $99$\% of mock bona fide images from DLC2021 as attacks.
Therefore, we decided to label the mock bona fide images from IJCB as bona fide and the mock bona fide from DLC2021 as attacks for the IJCB baseline model.
Table \ref{tab:id_card_datasets} shows an overview of the ID card datasets used for the experiments.

The corner detection model was trained on all $78,623$ images from the DLC2021 dataset, including ID cards and passports. 
For better generalization, we annotated $1,350$ additional images of ID cards from the other datasets in Table~\ref{tab:id_card_datasets}, merged them with the DLC2021 dataset and applied data augmentation during training.
We trained a new model, as the CHI and MEX datasets have many images that are already cropped and the corners sometimes lie outside the image.
Other pretrained models required the ID card to be fully visible.

\section{Results}\label{sec:results}

To evaluate PAD system performance, the equal error rate (EER) was used based on ISO/IEC 30107-3 \cite{isoiec}. 
Table \ref{tab:eer_datasets_pads} shows an overview of the PAD performance on all datasets, each merged with all bona fide images.

\begin{table}[h]
    \caption{EERs for all datasets and PADs}
    \centering
	\begin{tabular}{|l|c|c|c|c|}
		\hline
		\textbf{Dataset} & \textbf{PAD M1} & \textbf{PAD M2} & \textbf{IJCB Baseline} \\ \hline
		\textbf{CHI} & 27.50\% & 28.95\% & 20.51\% \\
		\hline
		\textbf{MEX} & 16.40\% & 19.52\% & 4.77\% \\
		\hline
		\textbf{DLC2021} & 23.83\% & 12.29\% & 6.42\% \\
		\hline
		\textbf{IJCB} & 17.18\% & 14.61\% & 4.38\% \\
		\hline
	\end{tabular}
	
	\label{tab:eer_datasets_pads}
\end{table}




To assess the effectiveness of the quality measures, the error versus discard characteristic (EDC) \cite{edc} was calculated.
An EDC shows the error rate of a PAD system when discarding a certain percentage of the lowest quality images according to a quality measure.
An EDC curve is a graph where the x-axis is the percentage of discarded images and the y-axis is an error rate, in our case the EER.
If the EDC curve shows a steep decrease, many images that lead to misclassifications by the PAD system were discarded.
If the EDC curve is flat, the quality measure does not have any influence on PAD performance.
If the EDC curve increases, images that were correctly classified by the PAD system were discarded. 
This characteristic was first proposed in \cite{edc_first}, as the error versus reject characteristic, but for the same reasons as in \cite{edc}, we use the term error versus discard characteristic to avoid confusing QA rejection with PAD rejection.

Since not all quality measures have a "higher is better" semantic, for some measures we removed the images with the lowest scores, and for others we removed the images with the highest scores. Depending on the PAD system or dataset, this semantic may be inverted or show no clear effect on PAD performance.

For a deeper investigation, we split the attack data into $13$ subsets ($12$ subsets for the IJCB baseline model), with one subset per attack per dataset including the mock bona fide images as attack.
From these subsets, all combinations of length $1,\ldots,13$ (or $1,\ldots,12$ for IJCB baseline) of the different subsets were created.
Each combination of subsets was merged into a single subset after equalizing the sizes of each subset to the maximum subset size of this combination and then further merged with the bona fide images.
Then, we calculated an EDC curve with EER as the error metric per subset, resulting in $8,191$ or $4,095$ EDC curves per measure and per PAD.
As we are interested in how much performance improves when discarding images, we divide each value of each EDC curve by the EER value without discarding anything.
We call this the relative error reduction versus discard characteristic (r$\Delta$EDC).
In r$\Delta$EDCs, a positive value corresponds to error reduction.
To keep the r$\Delta$EDC curves analogous to EDC curves, we flipped the y-axis, so that positive values are below the x-axis.

\begin{figure*}
    \centering
    \includesvg[width=1\textwidth]{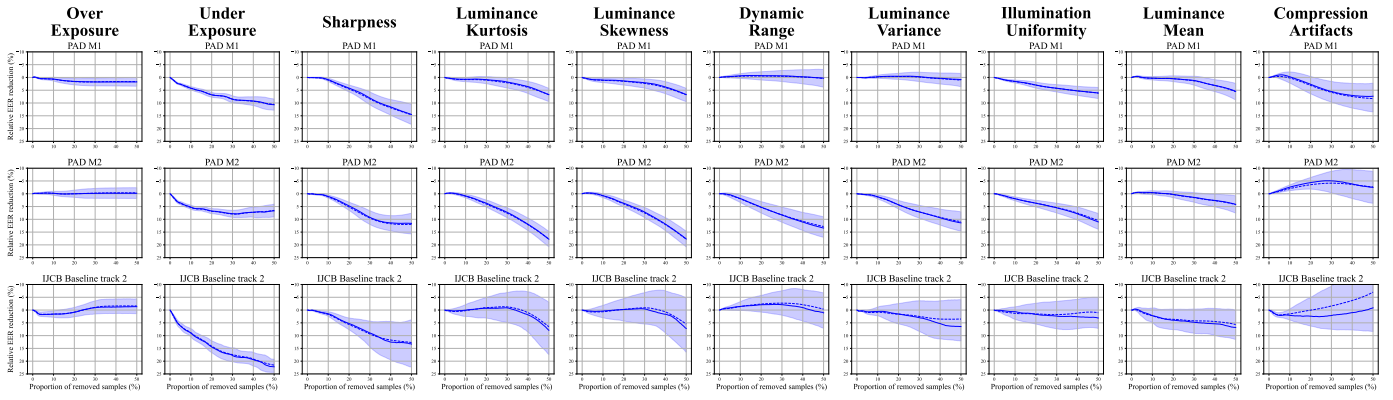}\vspace{-0.4cm}
    \caption{Aggregated r$\Delta$EDCs for each quality measure. Solid lines show the median r$\Delta$EDC, dashed lines show the mean r$\Delta$EDC and the shaded area shows the $25$th to $75$th percentile r$\Delta$EDC. }    
    \label{fig:edc}
\end{figure*}

To obtain a better overview, we aggregate the $8,191$ or $4,095$ r$\Delta$EDC curves per measure and per PAD by mean, median, and $25$th and $75$th percentile.
These aggregated r$\Delta$EDC curves are illustrated in Figure \ref{fig:edc} and mean improvements for removing $5$\%, $10$\% and $15$\% of the worst quality images according to each measure can be found in Table \ref{tab:mean_eer_reduction}.
Aggregated r$\Delta$EDC curves filtered by attack or dataset are available on \href{https://github.com/dasec/ID-Card-image-quality-on-OFIQ/tree/main/data/rDeltaEDC-curves}{GitHub}. 

\begin{table}[h]
    \caption{Mean relative EER reduction after removing the $5\%$, $10\%$, and $15\%$ worst samples according to each quality measure.}

	\centering
	\begin{tabular}{|c|l|c|c|c|}
		\hline
		\textbf{Measure} & \textbf{PAD} & \textbf{5\%} & \textbf{10\%} & \textbf{15\%} \\
		\hline
		\multirow{3}{*}{Over exposure} & PAD M1 & 0.51\% & 0.72\% & 1.27\% \\
		 & PAD M2 & -0.22\% & -0.17\% & 0.08\% \\
		 & IJCB Baseline & 1.64\% & 1.57\% & 1.43\% \\
		\hline
		\multirow{3}{*}{Under exposure} & PAD M1 & 2.69\% & 4.23\% & 5.35\% \\
		 & PAD M2 & 3.76\% & 5.36\% & 5.99\% \\
		 & IJCB Baseline & 6.13\% & 8.88\% & 11.46\% \\
		\hline
		\multirow{3}{*}{Sharpness} & PAD M1 & 0.14\% & 1.05\% & 2.80\% \\
		 & PAD M2 & 0.30\% & 1.22\% & 3.00\% \\
		 & IJCB Baseline & 0.63\% & 1.71\% & 3.68\% \\
		\hline
		\multirow{3}{*}{Luminance kurtosis} & PAD M1 & 0.68\% & 0.85\% & 0.84\% \\
		 & PAD M2 & -0.00\% & 0.96\% & 2.35\% \\
		 & IJCB Baseline & 0.56\% & 0.23\% & -0.37\% \\
		\hline
		\multirow{3}{*}{Luminance skewness} & PAD M1 & 0.90\% & 1.13\% & 1.25\% \\
		 & PAD M2 & -0.11\% & 0.82\% & 2.25\% \\
		 & IJCB Baseline & 0.74\% & 0.65\% & 0.18\% \\
		\hline
		\multirow{3}{*}{Dynamic range} & PAD M1 & -0.27\% & -0.46\% & -0.55\% \\
		 & PAD M2 & 0.59\% & 2.00\% & 3.68\% \\
		 & IJCB Baseline & -0.68\% & -1.36\% & -1.92\% \\
		\hline
		\multirow{3}{*}{Luminance variance} & PAD M1 & 0.05\% & -0.21\% & -0.34\% \\
		 & PAD M2 & 0.37\% & 1.26\% & 2.62\% \\
		 & IJCB Baseline & 0.91\% & 0.88\% & 1.20\% \\
		\hline
		\multirow{3}{*}{Illumination uniformity} & PAD M1 & 0.99\% & 1.62\% & 2.25\% \\
		 & PAD M2 & 0.88\% & 1.95\% & 2.87\% \\
		 & IJCB Baseline & 0.81\% & 1.29\% & 1.40\% \\
		\hline
		\multirow{3}{*}{Luminance mean} & PAD M1 & 0.02\% & 0.32\% & 0.41\% \\
		 & PAD M2 & -0.38\% & -0.40\% & -0.09\% \\
		 & IJCB Baseline & 0.94\% & 2.23\% & 3.35\% \\
		\hline
		\multirow{3}{*}{Compression artifacts} & PAD M1 & -0.37\% & 0.21\% & 1.54\% \\
		 & PAD M2 & -0.95\% & -2.01\% & -2.86\% \\
		 & IJCB Baseline & 1.93\% & 1.49\% & 0.75\% \\
		\hline
	\end{tabular}
	\label{tab:mean_eer_reduction}
\end{table}


First, we note that over exposure does not seem to have any influence on PAD performance, since the aggregated r$\Delta$EDC curves are mostly flat.
Under exposure, on the other hand, improves PAD performance by filtering out images with little to no dark pixels.
This may be because more dark pixels could indicate dark and clearly readable text on the ID card, which could help these PAD systems.
A strong influence on PAD performance can be seen for sharpness, but only after removing at least $10$\% of samples.
The improvement is expected, as a low sharpness score indicates a blurry image, which should make it difficult for the PAD system to extract relevant features.
Notably, luminance kurtosis and luminance skewness can also improve PAD performance, which contrasts with OFIQ, where they were discarded because they did not improve face detection performance.
Furthermore, dynamic range, luminance variance, and illumination uniformity appear to be relevant only for PAD M2.
Luminance mean shows a minor positive effect only for the IJCB baseline model, while compression artifacts show such an effect for PAD M1 after removing more than $10$\% of images.

Generally, PAD M2 is more strongly and consistently impacted by quality measures, while the IJCB baseline model exhibits high variance in measure effectiveness compared to the other two PAD systems.

%
%

\section{Conclusion and Future Work}\label{sec:conclusion}

In this work, image quality assessment methods for ID card images were developed by adapting OFIQ capture-related quality measures \cite{BSI_OFIQ1_0_2024}.
Of the 13 OFIQ capture-related measures, including three that were discarded in the final version of OFIQ, it was determined that ten of them can be applied to ID card images, with one (illumination uniformity) requiring a new algorithm, which was implemented by dividing the image into blocks and comparing the luminance median of each block.
The preprocessing pipeline includes corner detection using an hourglass network \cite{stacked_hourglass}, perspective normalization, and comprehensive foreground masking to ensure reliable quality measure computation and prevent biases.

The effectiveness of these quality measures was evaluated by analyzing their correlation with PAD performance using error versus discard characteristics.

The results suggest that assessing sharpness and under exposure has a clear positive influence on PAD performance across all three PAD systems and datasets, while over exposure does not have any effect on PAD performance.
The other measures (luminance kurtosis, luminance skewness, dynamic range, luminance variance, illumination uniformity, luminance mean and compression artifacts) have a moderate influence on some PAD systems.

Future work should focus on calibrating these measures for ID cards, as they were originally designed for face images.
New measures could also be developed specifically for ID cards, such as the degree to which the ID card is captured frontally or how strongly it is occluded.
During preprocessing, many pixels are masked out to prevent biases in the quality measures, but this means that a lot of information about the image is lost as well.
A shadow or bright reflection, which should lead to a worse quality score, may be completely masked out if it overlaps with text or a face image.
Developing more sophisticated methods to detect and score the strength of these artifacts could help improve quality assessment.
Afterwards, the native score values should be mapped to a scalar score in the range of $[0-100]$, similar to OFIQ, to project them all onto the same scale with a higher-is-better semantic for easier interpretability.
A final step would be to combine the individual quality measures into a single quality score for ID card images.

\section*{Acknowledgment}

This research work has been partially funded by the European Union (EU) under G.A. no. 101121280 (EINSTEIN) and CarMen (101168325), and the German Federal Ministry of Education and Research and the Hessian Ministry of Higher Education, Research, Science and the Arts within their joint support of the National Research Center for Applied Cybersecurity ATHENE.

\bibliographystyle{IEEEtran}
\bibliography{references}

\end{document}